\documentclass[letterpaper, 10 pt, journal, twoside]{IEEEtran}
\IEEEoverridecommandlockouts                              

\usepackage{graphicx} 
\graphicspath{{figs/}}
\usepackage{amsmath} 
\usepackage{amssymb}  
\usepackage{mathtools}
\usepackage{amsthm}
\usepackage{algorithmic}
\usepackage{multicol}

\usepackage{mleftright}
\usepackage[font=footnotesize]{caption}
\usepackage{placeins}
\usepackage{diagbox}
\usepackage{adjustbox}
\usepackage{verbatim}
\usepackage{cite}
\usepackage{wrapfig}
\usepackage{booktabs}
\usepackage{url}

\usepackage{todonotes} 

\def\Pr{\mathbf{Pr}}
\DeclareMathOperator*{\argmax}{arg\,max}

\usepackage[noline,linesnumbered]{algorithm2e}

\SetArgSty{textnormal} 

\def\Pr{\mathbf{Pr}}

\newtheorem{corollary}{Corollary}

\newcommand{\alg}{Power-UCT}

\newtheorem{innercustomgeneric}{\customgenericname}
\providecommand{\customgenericname}{}
\newcommand{\newcustomtheorem}[2]{%
  \newenvironment{#1}[1]
  {%
   \renewcommand\customgenericname{#2}%
   \renewcommand\theinnercustomgeneric{##1}%
   \innercustomgeneric
  }
  {\endinnercustomgeneric}
}

\newcustomtheorem{manualtheorem}{Theorem}
\newcustomtheorem{manuallemma}{Lemma}
\newcustomtheorem{manualassumption}{Assumption}

\title{\LARGE \bf
Monte-Carlo Robot Path Planning
}

\author{Tuan Dam$^{1}$, Georgia Chalvatzaki$^{1}$, Jan Peters$^{1}$ and Joni Pajarinen$^{1,2}$
\thanks{Manuscript received: February, 24, 2022; Revised May, 26, 2022; Accepted July, 23, 2022. This paper was recommended for publication by Editor Hanna Kurniawati upon evaluation of the Associate Editor and Reviewers' comments.
This project has received funding from the German Research Foundation project PA 3179/1-1 (ROBOLEAP), and the Emmy Noether Programme (\#448644653). (\textit{Corresponding author: Tuan Dam.})}
\thanks{$^{1}$Department of Computer Science, Technische Universit{\"a}t Darmstadt, Germany}%
\thanks{$^{2}$Department of Electrical Engineering and Automation, Aalto University, Finland}%
\thanks{\tt\footnotesize \{tuan,georgia,jan,joni\}@robot-learning.de}
\thanks{Digital Object Identifier (DOI): see top of this page.}
}

\begin{document}

\markboth{IEEE Robotics and Automation Letters. Preprint Version. Accepted July, 2022} {Dam \MakeLowercase{\textit{et al.}}: Monte-Carlo Robot Path Planning}

\maketitle

\begin{abstract}

Path planning is a crucial algorithmic approach for designing robot behaviors.
Sampling-based approaches, like rapidly exploring random trees (RRTs) or probabilistic roadmaps, are prominent algorithmic solutions for path planning problems. Despite its exponential convergence rate, RRT can only find suboptimal paths. On the other hand, $\textrm{RRT}^*$, a widely-used extension to RRT, guarantees probabilistic completeness for finding optimal paths but suffers in practice from slow convergence in complex environments. Furthermore, real-world robotic environments are
often partially observable or with poorly described dynamics, casting the application of $\textrm{RRT}^*$ in complex tasks suboptimal. This paper studies a novel algorithmic formulation of the popular Monte-Carlo tree search (MCTS) algorithm for robot path planning. Notably, we study Monte-Carlo Path Planning (MCPP) by analyzing and proving, on the one part, its exponential convergence rate to the optimal path in fully observable Markov decision processes (MDPs), and on the other part, its probabilistic completeness for finding feasible paths in partially observable MDPs (POMDPs) assuming limited distance observability (proof sketch). Our algorithmic contribution allows us to employ recently proposed variants of MCTS with different exploration strategies for robot path planning. Our experimental evaluations in simulated 2D and 3D environments with a 7 degrees of freedom (DOF) manipulator, as well as in a real-world robot path planning task, demonstrate the superiority of MCPP in POMDP tasks.
\end{abstract}


\section{Introduction}
\IEEEPARstart{R}{obot} path planning refers to the process of finding a sequence of configurations that lead a robot system from a starting configuration to a goal configuration without violating task constraints. Path planning is a crucial component in robotics~\cite{lavalle1998rapidly}, autonomous driving~\cite{ji2016path} and other domains such as surgical planning, computational biology, and molecules~\cite{latombe1999motion}.
In robotics, path planning is an integral tool for manipulation tasks with robotic manipulator arms~\cite{sahar1986planning, kim1985minimum, kunz2010real} and
mobile robots~\cite{zelinsky1993planning, alexopoulos1992path, zhang2018path}.

\begin{figure}[tbh!]
    \centering
    \includegraphics[scale=0.37]{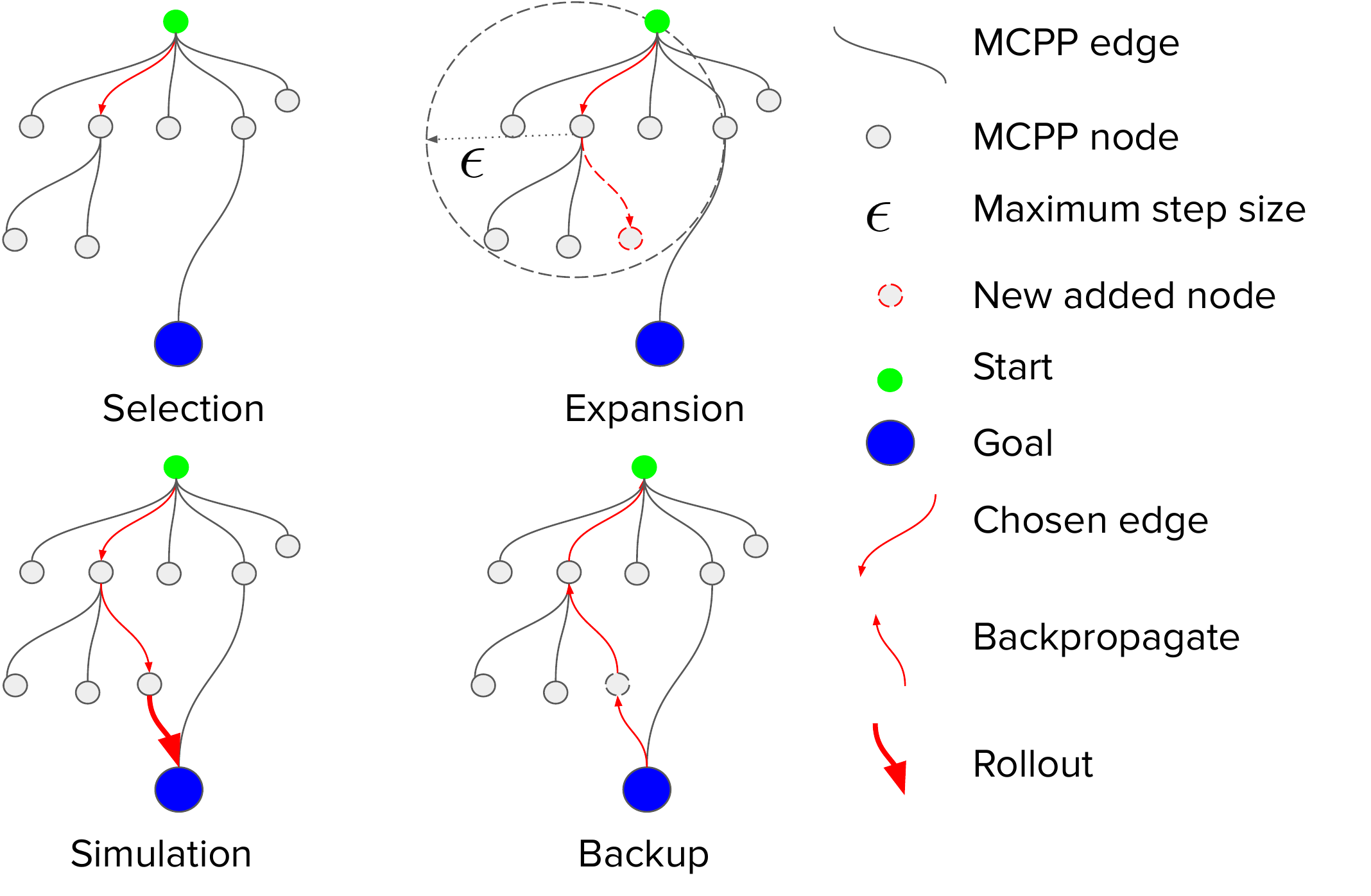}
    \vspace{-.5em}
    \caption{Four stages of MCPP planner to traverse from the initial position (in green color) to the goal position (in blue color).}
    \label{fig:mcts_mdp}
  \centering 
  \includegraphics[scale=0.38]{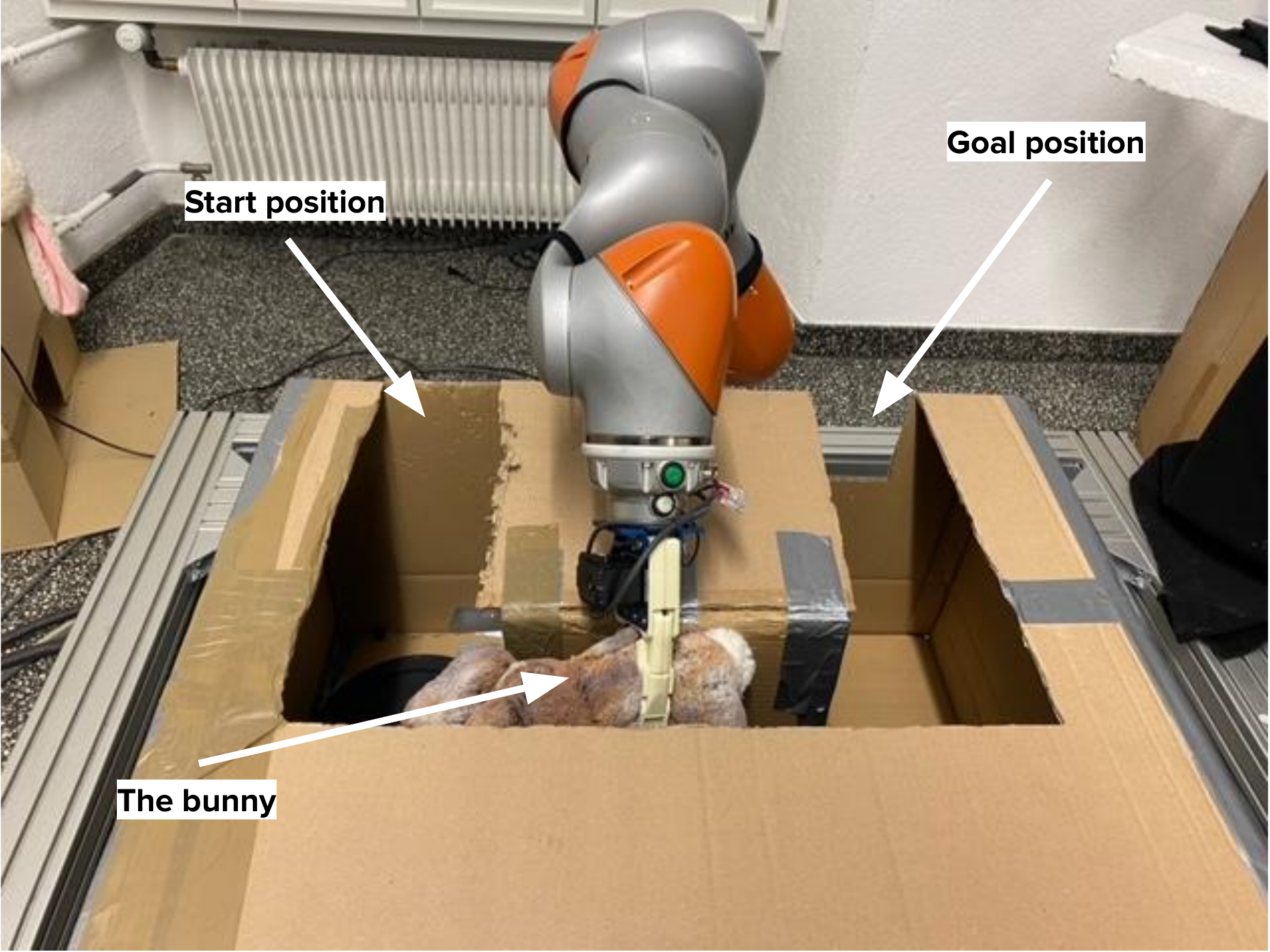}
  \vspace{-.5em}
  \caption{Demonstration of path planning using MCPP in a robotic disentangling task. A 7-DOF robotic KUKA arm tries to extract an object from the cardboard box through the hole in the back
  of the box. 
  The robot does not use any sensors except for proprioception, making the task partially observable.
  Therefore, the task requires advanced MCPP-based path planning that takes information gathering about the
  environment into account. We put a limit to prevent the robot arm to move the hand up, therefore, the robot arm has to find the path from the start position on the left side to the goal position on the right side inside the box.}  
  \label{fig:disentangling_robot}
  \vspace{-2.em}
\end{figure}

Due to the redundancy of robotic arms and the complexity and constraints of real-world tasks, sampling-based approaches yielded significant results~\cite{elbanhawi2014sampling,noreen2016optimal}. Among the different algorithmic contributions~\cite{persson2014sampling, lavalle1998rapidly,hsu1998finding, kavraki1996probabilistic}, $\textrm{RRT}^*$~\cite{kuffner2000rrt} is a widely used method that ensures finding the optimal path with probabilistic completeness guarantees~\cite{kleinbort2018probabilistic}. While $\textrm{RRT}^*$ is effective in solving path planning tasks in fully observable MDPs, real-world robotics applications are characterized by partial information, casting their settings into POMDP problems. In the real world, robots should make decisions based on information from laser sensors~\cite{ivanov2019mobile}, camera images~\cite{mezouar2002path}, and sensory feedback~\cite{lumelsky1987dynamic}, which generally contains noise, and subsequently makes it hard for planners such as $\textrm{RRT}^*$. Therefore, robot path planning under uncertainty~\cite{galceran2013survey, dadkhah2012survey,achtelik2014motion,kazemi2010path} has become one of the critical topics in the robotics community and remains an open research challenge.

This work proposes an algorithmic formulation to path planning problems based on the popular MCTS algorithm. 
We argue that the exploration-exploitation properties of MCTS algorithms are essential for robotic path planning in POMDPs, and they can outperform sampling-based planners like $\textrm{RRT}^*$ that greedily explore the state-space. To this end, we formulate an MCPP algorithmic framework that we analyze theoretically and provide proofs of convergence for the MDP and POMPD settings. In particular, when applying the upper confidence bounds for Trees (UCT) algorithm \cite{kocsis2006bandit}, we can guarantee the exponential convergence of MCPP to optimal paths in MDP problems. Crucially, we extend our theoretical analysis to prove the probabilistic completeness of MCPP in POMDP problems assuming limited distance observability. To the best of our knowledge, this is the first work to provide the theoretical analysis for MCTS in both MDP and POMDP robot path plannings.   
We continue by proposing different exploration strategies in MCPP for robotic path planning. In particular, we build on top of our prior work on power-mean UCT (Power-UCT)~\cite{dam2019generalized} and convex regularization with Tsallis Entropy Monte-Carlo Planning (TENTS)~\cite{dam2021convex}, integrating them in MCPP. We provide various experimental evaluations of MCPP, initially in MDP environments for completeness and thereafter in challenging POMDP tasks in 2D and 3D while planning with a 7-DOF robot arm. Moreover, we evaluate the different variants of MCPP against $\textrm{RRT}^*$ in a real-world POMDP experiment (see Fig. \ref{fig:disentangling_robot}), where the robot can only observe collisions in the box while planning to take out a bunny-toy. Our experimental results confirm that MCPP has a higher probability of solving POMDP path planning tasks with less planning time and requiring fewer samples than the baseline methods. We believe that our theoretical findings and empirical results will shed new light on robotic path planning in complex, partially observable tasks. 
To summarize, our \textit{contribution} is threefold:
\begin{itemize}
    \item we prove that MCPP enjoys exponential convergence in choosing the optimal path in MDP problems and has convergence guarantees to find a feasible path in POMDP environments with limited distance observability (for the POMDP case, we provide the proof sketch);
    \item using our theoretical insights, we propose an MCTS-based path planning framework that can incorporate different exploration strategies, such as our state-of-the-art methods, Power-UCT, and TENTS, into POMDP path planning problems;
    \item we provide empirical evaluations in simulation and real-world experiments that confirm our theoretical findings for the MCPP algorithmic framework to be a promising solution for planning in POMDP environments.
\end{itemize}
\section{Related work}\label{sec:related_work}

Probabilistic RoadMaps (PRMs) \cite{hsu1998finding} and RRTs \cite{kuffner2000rrt}  are fundamental approaches for sampling-based motion planning. $\textrm{RRT}^*$ improves over RRT by applying the rewiring technique to shorten the unnecessary traversing path. Moreover, $\textrm{RRT}^*$ has proven to guarantee probabilistic completeness for choosing the optimal path in MDP problems, but no convergence rate of $\textrm{RRT}^*$ has been studied so far.

There are several heuristic improvements over the state-of-the-art RRT and $\textrm{RRT}^*$. For example, A* is a sufficient heuristic path planning-based method for finding an optimal path given the graph representation of the environment. A*-RRT\cite{brunner2013hierarchical} integrates the benefit of the heuristic A* in RRT by sampling a new tree node using an A* path, and therefore improving the performance in terms of sample efficiency and cost compared to RRT. A*-$\textrm{RRT}^*$\cite{brunner2013hierarchical} combines A* with $\textrm{RRT}^*$ to improve the sample efficiency over $\textrm{RRT}^*$. Theta*-RRT \cite{palmieri2016rrt} considers Theta*, an any-angle discrete search method combined with RRT. Palmieri et al. \cite{palmieri2016rrt} prove that Theta*-RRT enjoys the probabilistic completeness of RRT and $\textrm{RRT}^*$, while finding shorter trajectories and plans significantly faster than baseline planners (RRT, A*-RRT, $\textrm{RRT}^*$, A*-$\textrm{RRT}^*$). Informed-$\textrm{RRT}^*$\cite{gammell2014informed} focuses the search on the ellipsoidal informed subset of the state-space of the initial running solution found by $\textrm{RRT}^*$.

Regarding applications of MCTS in path planning, Kim et al.~\cite{kim2020monte} proposes the use of Voronoi diagrams to discretize the action space and provides a regret-bound analysis for the sample efficiency, but the authors do not provide a convergence rate for goal reaching in the robot path planning setting. Sun et al.~\cite{sun2020stochastic} propose POMCP++, an improvement over POMCP~\cite{silver2010monte} to solve continuous observation problems. First, the authors propose using multiple particle samples from the current initial belief instead of a single particle sample of POMCP. 
Second, the authors handle the continuous observation space by proposing a new measurement sampling method. At each Q-node in the tree, POMCP++ either samples a new observation or chooses existing observations with some probability.
Experiments show that POMCP++ yields a significantly higher success rate and total reward. However, there is no actual convergence rate analysis in the robot path planning settings. Sunberg et al.~\cite{sunberg2018online} integrated the progressive widening technique in MCTS to discretize the continuous action and observation cases in POMDP settings and derived POMCPOW and POMCP-DPW (with double progressive widening). The authors further combined a weighted particle filter with progressive widening and showed the benefits over the baseline algorithm Determinized Sparse Partially Observable Tree (DESPOT) \cite{somani2013despot}.

Our work uses a simple uniform discretization of the action space for the MCTS algorithm in the context of robotic path planning. While our approach can apply the Voronoi diagram discretization of \cite{kim2020monte}, in this paper, we focus on the theoretical justification of our method and its comparison to sampling-based planners. We provide proofs of convergence for planning the optimal path to the goal in MDPs and a feasible path in POMDPs (with a proof sketch), which is not provided in \cite{sunberg2018online,sun2020stochastic}, but can also apply to them. Notably, we propose MCPP as a general MCTS-based framework for robotic path planning. MCPP can incorporate different exploration strategies \cite{dam2019generalized, dam2021convex} to continuous actions, adapting, subsequently, the convergence rates for MCPP. 

\section{Background}\label{sec:background}

\noindent\textbf{Markov Decision Process.} A finite-horizon MDP can be defined as a
$5$-tuple $\mathcal{M} = \langle \mathcal{S}, \mathcal{A},
\mathcal{R}, \mathcal{P}, \gamma \rangle$, where $\mathcal{S}$ is the
state-space, $\mathcal{A}$ is the finite action-space,
$\mathcal{R}: \mathcal{S} \times \mathcal{A} \times \mathcal{S} \to
\mathbb{R}$ is the reward function, $\mathcal{P}: \mathcal{S} \times
\mathcal{A} \to \mathcal{S}$ is the transition kernel, and $\gamma \in
[0, 1)$ is the discount factor. A policy $\pi \in \Pi: \mathcal{S}
  \times \mathcal{A} \to \mathbb{R}$ is a probability distribution of
  the event of executing an action $a$ in a state $s$.
Most sampling-based algorithms consider the environment as an MDP. Notably, in robot path planning problems, we
know the obstacle space so that when we sample a new vertex, we
can determine if the new sampled point lies in the free space or
not and then calculate the cost function.

\noindent\textbf{Partially Observable MDP.}
We consider a finite-horizon POMDP as a tuple $\mathcal{M} = \langle \mathcal{S},
\mathcal{O}, \mathcal{A}, \mathcal{R}, \mathcal{P}_s, \mathcal{P}_o,
\gamma \rangle$, where $\mathcal{S}$ is the state-space, $\mathcal{O}$
is the observation space, $\mathcal{A}$ is the finite action-space, $\mathcal{R}: \mathcal{S} \times \mathcal{A} \times \mathcal{S}
\to \mathbb{R}$ is the reward function, $\mathcal{P}_s: \mathcal{S}
\times \mathcal{A} \to \mathcal{S}$ is the state transition kernel,
$\mathcal{P}_o: \mathcal{O} \times \mathcal{A} \to \mathcal{S}$ is the
observation dynamics, and $\gamma \in [0, 1)$ is the discount
  factor. A policy $\pi \in \Pi: \mathcal{O} \times \mathcal{A} \to
  \mathbb{R}$ is a probability distribution of the event of executing
  an action $a$ in an observation $o$. In POMDP settings, the agent does not observe the full information of
the state of the environment, and the decisions are based only on observations. In general, the decision process can be made based either
on the history of all past actions and observations $h_t = \{a_0,
o_0, a_1, o_1,...,a_t, o_t\}$, or through the belief of the agent over the
state-space \cite{silver2010monte}.

\noindent\textbf{Monte-Carlo Tree Search.}
MCTS~\cite{baier2010power} combines tree search with Monte-Carlo sampling in order to build a tree, where states and actions are modeled as nodes and edges, respectively, to compute optimal decisions. The MCTS algorithm consists of a loop of four steps: \textit{\underline{Selection}}: start from the root node, interleave action selection and sample the next state (tree node) until a leaf node is reached; \textit{\underline{Expansion}}: expand the tree by
adding a new edge (action) to the leaf node and sampling the next state
(new leaf node); \textit{\underline{Simulation}}: rollout from the reached state
to the end of the episode using random actions or a heuristic;
\textit{\underline{Backup}}: update the nodes backward along the trajectory
starting from the end of the episode until the root node according to
the rewards collected.

UCT~\cite{kocsis2006bandit,kocsis2006improved} is an extension of the well-known UCB1~\cite{auer2002finite} multi-armed bandit algorithm. UCB1 chooses the arm (action $a$) using
\vspace{-0.5em}
\small
\begin{flalign}
a = \argmax_{i \in \{1...K\}} \overline{Q}_{i, T_i(n-1)} + C\sqrt{\frac{\log n}{T_i(n-1)}}.
\label{eq:UCB1}
\end{flalign}
\normalsize
where $T_i(n) = \sum^n_{t=1} \textbf{1} \{t=i\} $ is the number of
times arm $i$ is played up to time $n$, and $\overline{Q}_{i, T_i(n-1)}$
is the average reward of arm $i$ up to time $n-1$ and $C
= \sqrt{2}$ is an exploration constant. 
In UCT, the value of each node is backed up recursively from the leaf node to the root node as averaging over the child nodes.
At each action selection step in MCTS, each arm in the tree is chosen as the maximum value of nodes in the current non-stationary multi-armed bandit setup, as in (\ref{eq:UCB1}).
UCT ensures the asymptotic convergence of choosing the optimal arm at the root node~\cite{kocsis2006improved}.

Power-UCT \cite{dam2019generalized}, an improvement over UCT,  solves the problem of the underestimation of the average mean and the max-backup
operators in MCTS by proposing the use of power mean as the backup
operator. Power-UCT has a polynomial convergence rate for choosing the optimal action at the root node. TENTS~\cite{dam2021convex} is derived as a result of Tsallis entropy
regularization in MCTS. TENTS has an exponential convergence rate at the root node, which is faster than Power-UCT and UCT. TENTS has a lower value error and smaller regret bound at the root node compared to other regularization approaches.

\section{Problem formulation}\label{sec:problem_formulation}
Let us define the robot path planning problem, both
for MDPs and POMDPs.
Let $\mathcal{X} = (a,b)^d$ be the configuration space of the robot, where $a,b \in
\mathcal{R}$ are joint limits in configuration space, with $a < b$, and $d \in \mathcal{N}, d > 0$ denoted the robot's DOF. Let's define $\mathcal{X}_{\textrm{OBS}}$ as the obstacles region
and $\mathcal{X} \backslash \mathcal{X}_{\textrm{OBS}}$ the open set, and the obstacle-free space as $\mathcal{X}_{\textrm{FREE}} = cl(\mathcal{X} \backslash \mathcal{X}_{\textrm{OBS}})$, where $cl(\cdot)$ denotes the closure of a set. The initial condition, or start region, $\mathbf{x}_{\textrm{INIT}}$ is an element of $\mathcal{X}_{\textrm{FREE}}$, and the goal region $\mathbf{x}_{\textrm{GOAL}}$ is an open subset of $\mathcal{X}_{\textrm{FREE}}$. A path planning problem is defined by the triplet $(\mathcal{X}_{\textrm{FREE}}, \mathbf{x}_{\textrm{INIT}}, \mathbf{x}_{\textrm{GOAL}})$.  A trajectory is defined as the map $\tau:[0,T] \rightarrow \mathcal{X}_{\textrm{FREE}}$, where  $\tau(0) = \mathbf{x}_{\textrm{INIT}}$, $\tau(T) = \mathbf{x}_{\textrm{GOAL}}$. Let's define a function $\sigma: \mathcal{X} \times \mathcal{X} \rightarrow \mathbb{R}$ as the cost function for moving the robot from the configuration point $\mathbf{x}_i$ to $\mathbf{x}_j$, where  $\mathbf{x}_i, \mathbf{x}_j \in \mathcal{X}$. A solution to such a problem is a trajectory that moves the robot from the initial region to the goal region, while avoiding collisions with obstacles and having minimum cost. 

\noindent \textit{\underline{Fully observable problem.}} Here, we assume that we know the state of the environment, i.e.,  we know the $\mathcal{X}_{\textrm{OBS}}$ space and $\mathcal{X}_{\textrm{FREE}}$ regions. Whenever a new point is sampled in the configuration space $\mathcal{X} = (a,b)^d$, we can measure the cost and determine if the point is inside the free space or not.

\noindent \textit{\underline{Partially observable problem.}} In this setting, we assume that the environment is partially observable, i.e., we only know the start position and the goal position. We do not observe the full state but only observations of the environment and progressively build a belief about the environment's state from observations.

\section{Monte-Carlo path planning}\label{sec:analysis}

We wish to transform MCTS into a sampling-based method for solving robot path planning problems when applicable. We build our proposed MCPP approach starting from the UCT algorithm. MCPP and UCT share similar ways of selecting nodes to traverse and back up the value of nodes in the tree. However, we need to make several algorithmic choices to do path planning with UCT. First, we draw an $\epsilon$-ball to limit the maximum distance that the robot can move from the current configuration point. Second, we perform uniform sampling of the configuration points inside the $\epsilon$-ball to discretize the continuous actions in the MDP. Third, we investigate different exploration strategies for MCPP, like in the PowerUCT and TENTS algorithms. We provide a proof of the exponential convergence rate of finding the optimal path in MDPs. Moreover, we connect this result to Power-UCT and TENTS and derive their respective convergence rates for path planning. In POMDPs, we provide a probabilistic completeness guarantee for finding the feasible path to the goal with limited distance observations.
\subsection{Fully observable environment}
In an MDP, the agent knows the full state of the environment. Let us define the start position as
$\mathbf{x}_{\textrm{INIT}}$ and the goal position as $\mathbf{x}_{\textrm{GOAL}}$. The cost function is the Euclidean
distance $d(\mathbf{x},\mathbf{y})$ between two points $\mathbf{x}$ and $\mathbf{y}$. We want to minimize the
total cost, that is, the total distance traveled from the start to the goal position using MCPP.
\begin{figure}[t!]
  \centering  
  \includegraphics[scale=0.33]{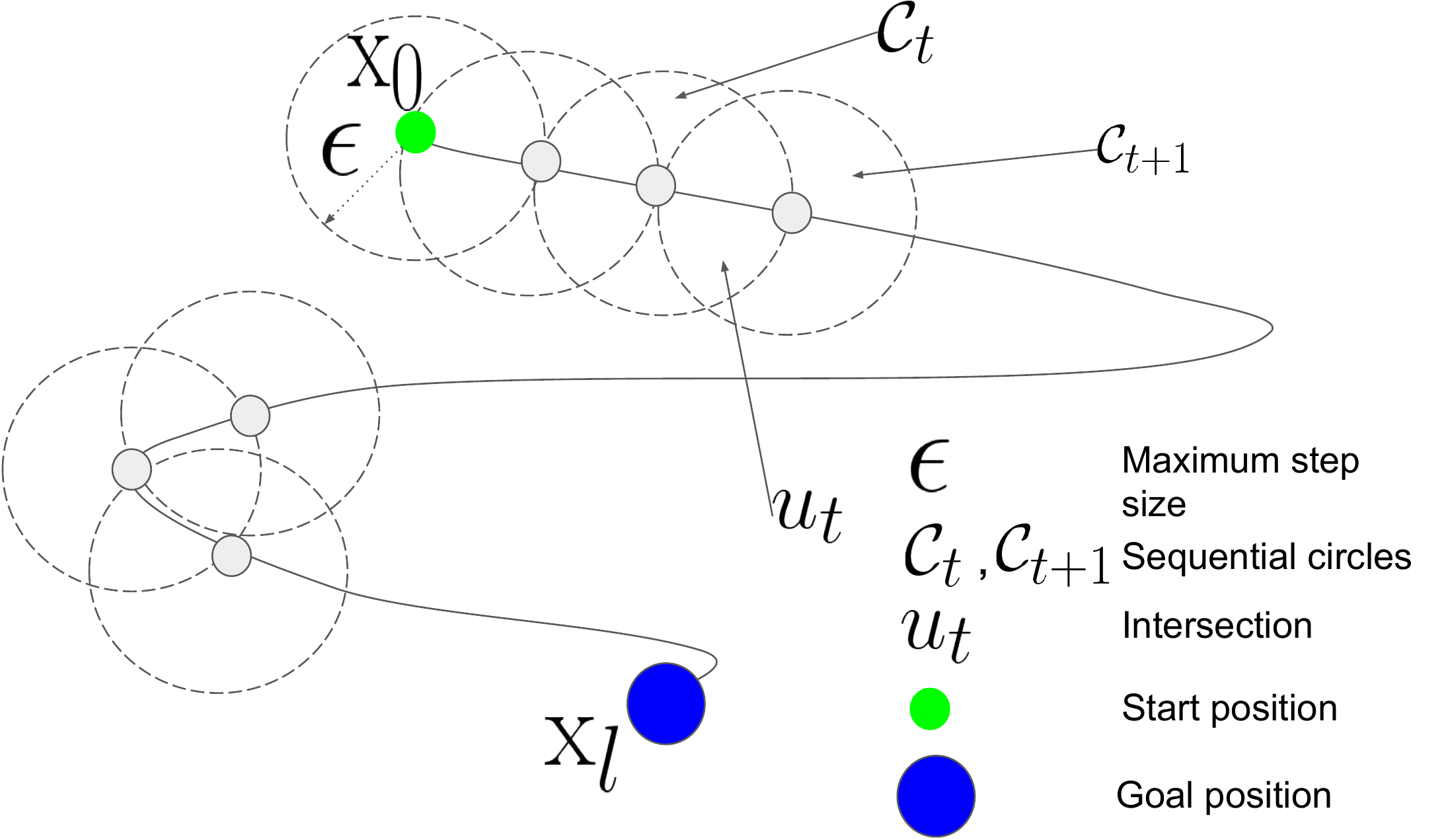}
  \caption{2-D sketch of the proof for exponential convergence of MCPP to the optimal path in MDPs.
  The MDP proof relies on showing that MCPP convergences exponentially to a path starting from $\mathbf{x}_0$ and ending at $\mathbf{x}_l$ while the agent stays inside a tube composed of a sequence of spheres with a radius of $\epsilon$.}
  \label{fig:mcts_path_sketch}
    \vspace{-2em}
\end{figure}
As shown in Fig.~\ref{fig:mcts_mdp}, at each node of the tree, starting from the root node, actions are generated by uniformly sampling random
points in the $\epsilon$-ball distance from the current node. 

The Algorithm~\ref{alg:mcts_path_planning} provides the pseudocode of the MCPP method in the MDP case.
The $\texttt{MainLoop}$ procedure is the main loop of the algorithm. The algorithm stops when the $\mathbf{x}_{\textrm{GOAL}}$ position is reached.
The algorithm follows the four basic steps of a regular MCTS method. First, at the \textit{\underline{Selection}} step, we determine the next node to traverse in the tree by selecting the action as in the $\texttt{SelectAction}$ procedure. Here, an action is selected based on the UCB algorithm. Note that when we implement Power-UCT~\cite{dam2019generalized} we also use UCB, while TENTS\cite{dam2021convex} uses stochastic Tsallis entropy regularization for the action sampling. 
Second, at the \textit{\underline{Expansion}} step, $|A|$ number of actions are generated by uniformly sampling inside the circle $\mathcal{C}(s,\epsilon)$ as shown in the $\texttt{Expand}$ procedure. When we reach the leaf node, a new node is created and is added to the MCTS tree. Third, at the \textit{\underline{Simulation}} step, as shown in the $\texttt{Rollout}$ procedure, the value function of the current node $s$ is calculated as the distance from that node $s$ to the goal position. Finally, at the \textit{\underline{Backup}} step, the return value is backpropagated in the two procedures $\texttt{SimulateV, SimulateQ}$.
\subsection{Partially observable environment}
Under partial observablity, the agent does not
observe the full state of the environment, but has only access to possibly noisy observations. 
The MCPP planner makes decisions based on the current belief
of the agent over the state of the environment. Therefore, our
approach in POMDP will be the same as in MDP, except for the fact that we do the planning in the belief space. 
The other choice is that MCPP planner can make decisions over the history of actions and observation
as if it has some sufficient statistic~\cite{Smith2005PointBasedPA, braziunas2003pomdp}.
\SetKwProg{Fn}{}{}{}
\begin{algorithm*}
\small
  \caption{Pseudocode of MCPP.}
  \label{alg:mcts_path_planning}
  \begin{multicols}{2}
  \SetKwFunction{SelectAction}{SelectAction}
  \SetKwFunction{Search}{Search}
  \SetKwFunction{PartialObservableAStar}{PartialObservableAStar}
  \SetKwFunction{Expand}{Expand}
  \SetKwFunction{Rollout}{Rollout}
  \SetKwFunction{SimulateV}{SimulateV}
  \SetKwFunction{SimulateQ}{SimulateQ}
  \SetKwFunction{MainLoop}{MainLoop}
  $|A|$: number of actions.\\
  $N(s)$: number of simulations of V\_Node. Default is $0$.\\
  $n(s,a)$: number of simulations of Q\_Node. Default is $0$.\\
  $r, r(s,a)$: intermediate rewards defined as the distance between two nodes.\\
  $V(s)$: Value of V\_Node at state $s$. Default is $0$.\\
  $Q(s,a)$: Value of Q\_Node. Default is $0$.\\
  $\gamma$: discount factor. Default is $1$.\\  
  \BlankLine
  \Fn{R = \Rollout{$s$}}{
    $R = $ Distance from the current node s to the goal position.\\
    \Return $R$
  }
  \BlankLine
  \Fn{a = \SelectAction{$s$}}{
    \Return $\argmax_{a} Q(s, a) + C\sqrt{\frac{\log N(s)}{n(s,a)}}$\\
  }
  \BlankLine
  \Fn{a = \Search{$s$}}{
    \While{Time remaining}{
      \SimulateV($s$)\\
    }
    \Return $\argmax_a Q(s,a)$\\
  }
  \BlankLine
  \Fn{R = \Expand{$s$}} {
    Generate $|A|$ actions for the current node $s$ by randomly sampling $|A|$ via-points inside
    the circle $\mathcal{C}(s,\epsilon)$\\
    $discountedReward = \Rollout(s)$\\
    \Return $discountedReward$\\
  }
  \BlankLine
  \Fn{\SimulateV{$s$}} {
    $a = $\SelectAction($s$)\\
    \SimulateQ($s,a$)\\
    $N(s) = N(s) + 1$\\
    $V(s) = \big (\sum_{a} \frac{n(s,a)}{N(s)} Q(s,a) \big )$\\
  }
  \BlankLine
  \Fn{\SimulateQ{$s,a$}}{
    $(s',r) \sim \tau(s,a)$\\
    \uIf{$V(s') \text{ not expanded}$} {
    $r = r + \gamma . \Expand(s'$)\\
    } \Else {
        \SimulateV($s'$)\\
    }
    $r(s,a) = r(s,a) + r$\\
    $n(s,a) = n(s,a) + 1$\\
    {$Q(s,a) = \frac{(\sum_a r_{s,a}) + \gamma.\sum_{s'} N(s').V(s')}{n(s,a)}$}\\
    where $V(s')$ is the value function of the next node by action $a$ from the current $Q(s,a)$ node \\
  }
  \BlankLine
  \Fn{\MainLoop}{
    \While{Xgoal is reached}{
        $a = \Search{s}$\\
        $Execute(a)$ in real Robot
    }
  }
  \BlankLine
  \end{multicols}
\end{algorithm*}
\normalsize
\subsection{Theoretical analysis}
In this section, we prove that MCPP ensures an exponential convergence rate for finding the optimal path from the start position to the goal position in an MDP environment. In a POMDP setting, we prove that there is a high probability that MCPP can find the path to the goal position. 
\subsubsection{MDP}
First, we make the following assumption.
\begin{manualassumption}{1}\label{assumpt_1}
There exists an optimal path from the start position $\mathbf{x}_{\textrm{INIT}}$ to the goal
position $\mathbf{x}_{\textrm{GOAL}}$ with $\delta$ clearance (minimum distance to an obstacle).
\end{manualassumption}
Based on this assumption, we derive a theorem for the convergence
rate of finding the optimal path using MCPP:
\begin{manualtheorem} {1}
The probability that MCPP fails to find the optimal path from $\mathbf{x}_{\textrm{INIT}}$ to
$\mathbf{x}_{\textrm{GOAL}}$ after $n$ simulations is at most $a e^{-bf(n)n}$, for some constants
$a, b \in R_{>0}$.
\vspace{-0.5em}
\end{manualtheorem}
\begin{proof}
Let us consider all feasible paths from the start position ($\mathbf{x}_{\textrm{INIT}}$) to the goal position ($\mathbf{x}_{\textrm{GOAL}}$).
We will prove that MCPP ensures probabilistic completeness of finding the shortest path from ($\mathbf{x}_{\textrm{INIT}}$) to ($\mathbf{x}_{\textrm{GOAL}}$). We will further prove that the failure probability of finding the shortest path decays exponentially for an infinite number of samples. 

We choose a ball with radius $\epsilon = \delta$, where $\delta$ is the clearance of the shortest path $\tau^*$.
Along the path $\tau^*$, we define a set of $l+1$ circles with the radius $\epsilon$ and the center $\textrm{x}_t (t = 0...l) \in \mathcal{X}_{\textrm{FREE}}$. Here $\textrm{x}_0$ = $\mathbf{x}_{\textrm{INIT}}$ and $\textrm{x}_l$ = $\mathbf{x}_{\textrm{GOAL}}$, as shown in Fig.~\ref{fig:mcts_path_sketch}.
We define each circle $\mathcal{C}_t = (\textrm{x}_t, \epsilon), t = {0,1,...l}$.
We define the intersection set $u_t = \mathcal{C}_t \cap \mathcal{C}_{t+1}$. Let $p$ be the probability that MCPP can move from $\mathcal{C}_t$ to $\mathcal{C}_{t+1}$.
Consider starting from planning node $\textrm{x}_t$, which is the center of the circle $\mathcal{C}_t$.
If the next planning node $\textrm{x}_{t+1}$ lies in the circle $\mathcal{C}_{t+1}$, it has to lie inside the intersection $u_t$, and we can see that $p < 1/2$.
For the robot to travel from $\textrm{x}_0$ to $\textrm{x}_l$, it has to use at least $l$ MCPP vertices.
Let the probability that MCPP chooses the best action (action with smallest cost) be $f(n)$.
Therefore, the probability of MCPP of taking an optimal action that also lies inside $u_t$ is $f(n)p$.
The failure probability that MCPP cannot find the shortest path $\tau^*$ from $\textrm{x}_0$ to $\textrm{x}_l$ is $\Pr(X_n < l)$ where $X_n$ is the number of circles $\mathcal{C}_t, t \in {1,2,...l}$ which are connected by vertices, that is, for an optimal path, all circles need to be connected by vertices.
To calculate $X_n$, the initial value of $X_n$ is zero. We will incrementally
increase $X_n$ 
by one when a new circle along the optimal path is connected with a new vertex.
When $X_n$ is equal to or greater than $l$, we, then, have found the optimal path.
Let us upper bound the failure probability $\Pr(X_n < l)$ by first upper bounding $\Pr(X_n = h)$, as
\vspace{-0.5em}
\begin{flalign}
\Pr(X_n = h) &\leq \binom n h (f_{0}p)^h (1-f_{0}p)^{n-h} \label{fail_bound}\\
&\leq \binom n h (f(n)p)^h (1-f_{0}p)^{n-h}, \nonumber
\vspace{-0.5em}
\end{flalign}
\normalsize
where $(f(n)p)^h$ is the upper bound probability of having $h$
circles connected by vertices and $\binom n h$ makes sure there is at least one consecutive sequence of connected circles. $f_{0}$ is the initial probability of choosing the optimal action. 
\eqref{fail_bound} can be explained as $H(x) = x^{h} (1-x)^{n-h}$ is a decreasing function.
This yields an upper bound for $\Pr(X_n = h)$:
\vspace{-0.5em}
\small
\begin{flalign}
\Pr(X_n = h) &\leq \binom n h (f(n)p)^h (1-f_{0}p)^{n-h} \nonumber\\
&\leq \binom n h (f(n)p)^h (1-\alpha f(n)p)^{n-h} \nonumber
\end{flalign}
\normalsize
where $f_{0} = \alpha$ is a constant and $f(n) \leq 1$ so that $1-f_{0}p \leq 1-\alpha f(n)p$.
The probability of failing to find the optimal path is then
\vspace{-1em}
\small
\begin{flalign}
\Pr(X_n < l) &= \sum_{h=0}^{l-1} Pr(X_n = h) \nonumber\\
&\leq \sum^{l-1}_{h = 0} \binom n h (f(n)p)^h (1-\alpha f(n)p)^{n-h} \nonumber
\end{flalign}
\vspace{-1em}
\small
\begin{flalign}
&\leq \sum^{l-1}_{h = 0} \binom n{l-1} (f(n)p)^h (1-\alpha f(n)p)^{n-h}\; (\text{as }l << n) \nonumber
\end{flalign}
\vspace{-1em}
\small
\begin{flalign}
&\leq \binom n{l-1} \sum^{l-1}_{h = 0} (1-\alpha f(n)p)^{n}\; \nonumber\\
&(\text{as }f(n)p < 1/2 \text{ so that }f(n)p < 1 - f(n)p < 1 - \alpha f(n)p) \nonumber \\
&\leq \binom n{l-1} \sum^{l-1}_{h = 0} (e^{-\alpha f(n)p})^{n} = \binom n{l-1} l e^{-\alpha f(n) p n} \notag\\ &(\text{as }1-\alpha f(n)p <= e^{-\alpha f(n)p}) \nonumber\\
&=\frac{\prod^{n}_{i=n-l}i}{(n-1)!}le^{-\alpha f(n) p n} \leq \frac{l}{(l-1)!}n^l e^{-\alpha f(n) p n} \leq  a e^{-bf(n) p n} \nonumber
\end{flalign}
\normalsize
The provided convergence rate proves that the MCPP algorithm is probabilistically complete and converges to the optimal path exponentially.
\end{proof}

Let us define $g(t)$ as the failure probability of finding the optimal path
from $\mathbf{x}_{\textrm{INIT}}$ to $\mathbf{x}_{\textrm{GOAL}}$ after $t$ time steps. We derive the following
corollaries:\\
With Power-UCT, $f_{\text{Power-UCT}} = 1 - (\frac{1}{t})^{\alpha})t$.
The probability that the MCPP using Power-UCT fails to find the path from $\textrm{x}_0$ to $\textrm{x}_l$ is as follows.
\begin{corollary}Power-UCT:\\
$g_{\text{Power-UCT}}(t) = ae^{-b(1 - (\frac{1}{t})^{\alpha})t}, \text{where } 0 < \alpha < 1, a, b \in \mathbb{R}_{>0}$
\end{corollary}
\noindent With TENTS $f_{\text{TENTS}} = 1 - ct\exp\{-\frac{t}{\hat{c}(\log{t}^3))t}\}$.
The probability that MCPP using TENTS fails to find the path from $\textrm{x}_0$ to $\textrm{x}_l$ is
\begin{corollary}TENTS:\\ 
$g_{\text{TENTS}}(t) = ae^{-b(1 - ct\exp\{-\frac{t}{\hat{c}(\log{t}^3))t}\}}, \text{where } a, b, c, \hat{c} \in \mathbb{R}_{>0}$
\end{corollary}

\noindent The results show that MCPP-TENTS robot path planning converges faster
compared to MCPP-Power-UCT.
\subsubsection{POMDP}
First, we make the following assumption.
\begin{manualassumption}{2}\label{assumpt_2}
The agent observes the environment only up to $\gamma$ distance.
\end{manualassumption}

This assumption is reasonable in many robotic settings, e.g., for mobile robotics. Based on this assumption and Assumption~\ref{assumpt_1}, we derive a theorem to show that with high probability, the MCPP algorithm can find the feasible path to the goal position in a POMDP environment: 
\begin{manualtheorem} {2}
In POMDP environments with limited distance observability, MCPP will find a path from the start position $\mathbf{x}_{\textrm{INIT}}$ to the goal position $\mathbf{x}_{\textrm{GOAL}}$ with high probability. 
\vspace{-.5em}
\end{manualtheorem}

\begin{figure}[t!]
  \centering 
  \includegraphics[scale=0.25]{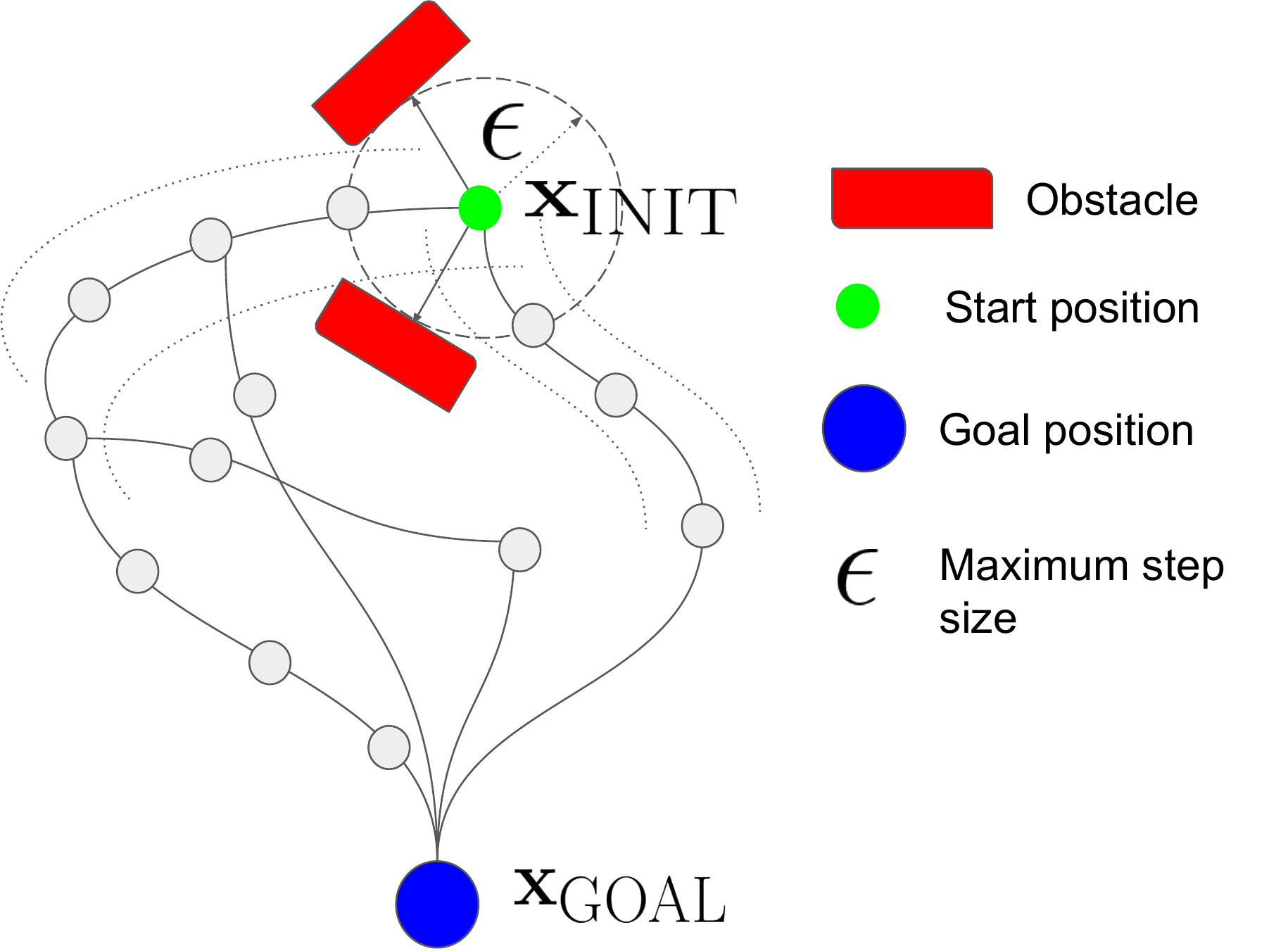}
  \caption{Sketch of how to generate paths for MCPP algorithm from $\mathbf{x}_{\textrm{INIT}}$ to $\mathbf{x}_{\textrm{GOAL}}$ positions with minimum number of via-points in POMDP environments.}  
  \label{fig:pomcp_path_sketch}
  \vspace{-2em}
\end{figure}
\begin{proof}
We assume that there is a finite number of feasible paths ($\tau_1, \tau_2, ... \tau_K$) to go from the start position $\mathbf{x}_{\textrm{INIT}}$ (or $\mathbf{x}_0$) to the goal position $\mathbf{x}_{\textrm{GOAL}}$. Each feasible path $\tau_i$ has at least $\delta_i$ clearance from obstacles. We choose $\epsilon = \min \{ \delta_1, \delta_2,...\delta_K, \gamma \}$. $\gamma$ is the observation distance defined in Assumption \ref{assumpt_2}.

\noindent Along each path $\tau_i$, let us define a set of circles $\mathcal{C}_i = (\mathbf{x}_i, \epsilon), i = {0,1,...l_{\tau_i}}. $ as shown in Fig.~\ref{fig:mcts_path_sketch}. Let us define $\mathbf{C}$ as the set of all circles (along all the feasible paths that we define). We assume that if the agent collides, the agent moves back to the last planning point and will not go to the direction of the obstacle again with high probability. We define that the probability $p_{\text{collision}} \rightarrow 0$ when the time step $t \rightarrow \infty$.
We prove that with high probability, the agent can find the path from the start position $\mathbf{x}_{\textrm{INIT}}$ to the goal position $\mathbf{x}_{\textrm{GOAL}}$.

The proof is derived by induction. From the start position $\mathbf{x}_0$, there is a finite number of circles $\mathcal{C}_i \in \mathbf{C}$ as the next feasible region that the MCPP planner can sample as the next node in the tree (MCPP samples the next planning point inside the $\epsilon$-ball distance). Because the probability of colliding again is $p_{\text{collision}} \rightarrow 0$ when the time step $t \rightarrow \infty$, and the MCPP objective is to minimize the cost to go to the goal position. When we increase the number of samples, the next planning node will lie inside the circle that contains the optimal path. Therefore, with high probability $1 - p_{\text{collision}}$ the next MCPP node will be inside one of the circles $\mathcal{C}_i$.
Assume now that the agent is inside the circle $\mathcal{C}_i$. Using the same induction, there is high probability $1 - p_{\text{collision}}$ that the next MCPP node will be inside one of the next circle $\mathcal{C}_{i+1}$. $\mathcal{C}_{i+1} \in \mathbf{C}$. Since the number of circles is finite, the agent will get to the goal region after a certain number of time steps with high probability, concluding the proof.
\end{proof}
\vspace{-0.5 em}
\section{Experiments}\label{sec:experiment}
In this section, we evaluate the performance MCPP in challenging POMDP environments. In MCPP, we apply the two recent advanced improvement techniques in MCTS, \alg~and TENTS, along with the baseline MCTS method, UCT.
We compare our new robot path planning methods against the baseline sampling-based method $\textrm{RRT}^*$, and a state-of-the-art continuous action POMDP solver POMCP-DPW~\cite{sunberg2018online}. 
In simulation, we also compared against two different
simple heuristic methods. The first method puts a ball around the agent to sample the next point. We use the same step size (the ball's diameter) and the same number of samples as MCPP and RRT*. 
In the second heuristic, we use an $\epsilon$-greedy probability ($1\%$) to sample the goal position or otherwise the next node, similarly to random node sampling in RRT*. We do not put any restrictions on the step size to sample the next node.

The POMDP setting of our experiment is the same as in~\cite{pajarinen2020probabilistic}. Similarly to the path planning definition in Sec.~\ref{sec:problem_formulation}, the state space $\mathcal{X}$ consists of configuration space coordinates such as robot joint angles.
The action space $\mathcal{A}$ is identical to the state space, consisting of target configuration space coordinates defining where to move the robot. We use linear interpolation to move the robot from the current configuration to the next one. The observation is a configuration in collision with an obstacle in obstacle space $\mathcal{X}_{\textrm{OBS}}$, or the goal configuration $\mathbf{x}_{\textrm{GOAL}}$, when the robot reaches the goal.
We assume static obstacles and deterministic transition and observation probabilities $\mathcal{P}_s$ and $\mathcal{P}_o$. We define the reward function as a success pseudo-probability along the path from one configuration ($\mathrm{x}_1$) to another ($\mathrm{x}_2$): $\mathcal{R}(\mathrm{x}_1, \mathrm{x}_2) = P_\textrm{SUCCESS}(\mathrm{x}_1, \mathrm{x}_2)$ where $P_\textrm{SUCCESS}(\mathrm{x}_1, \mathrm{x}_2)$ is defined in \cite{pajarinen2020probabilistic}. We set the discount factor $\gamma = 1$ and limit the planning horizon. As in \cite{pajarinen2020probabilistic}, to approximate the belief over states, based on prior collisions, we compute a probabilistic map that assigns a probability of colliding to any given position in the environment. The belief distribution is able to represent multi-modal and asymmetric belief distributions (see Fig.~2 in \cite{pajarinen2020probabilistic}). Initially, we assume a non-zero probability of colliding at any location on the map. After each collision, we update the map by assigning a failure probability that takes into account the collision coordinates and the movement direction (see Fig.~2 in \cite{pajarinen2020probabilistic}).

Following the previous POMDP definition, we evaluate the methods in simulation in two 7-DOF configuration space POMDP tasks with 2D and 3D task spaces. Finally, we compare MCPP to RRT* in a real robot POMDP disentangling application, similar to the one described in~\cite{pajarinen2020probabilistic}.

\subsection{Experimental evaluation in simulation}

\begin{figure}[t!]
  \centering  
  \includegraphics[scale=0.33]{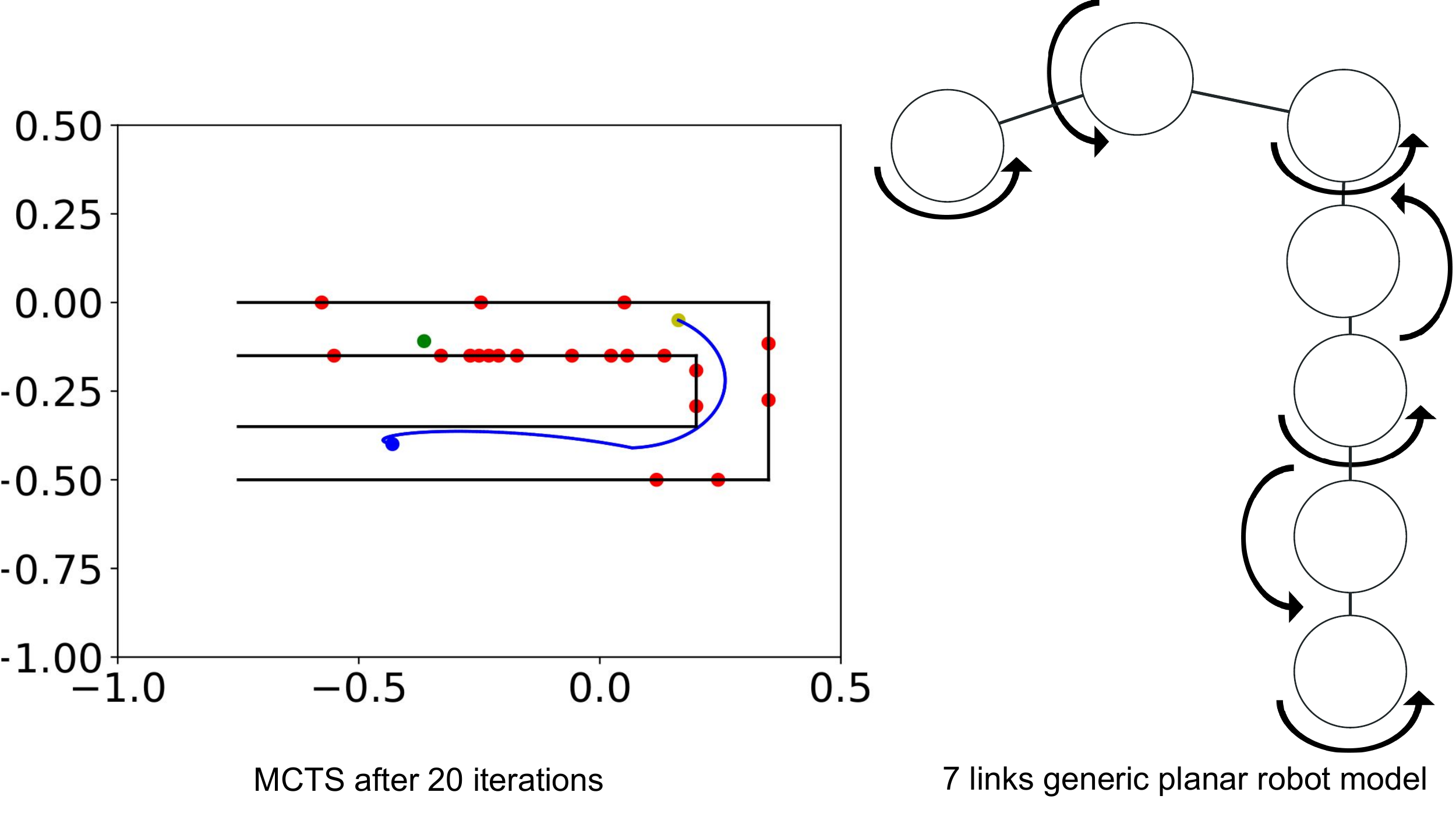}
  \vspace{-0.5em}
  \caption{U-Shape 2D POMDP. Green point is the start position. Blue point is the goal position. Red points are the collisions. The figure shows a success case using MCPP, where the blue line depicts the 2D trajectory of the end effector.
  Note that in all the 2D experiments we plan in the configuration space using a 7-DOF planar robot arm model illustrated on the right.}  
  \label{fig:2d_u_shape}

  \centering  
  \includegraphics[scale=0.35]{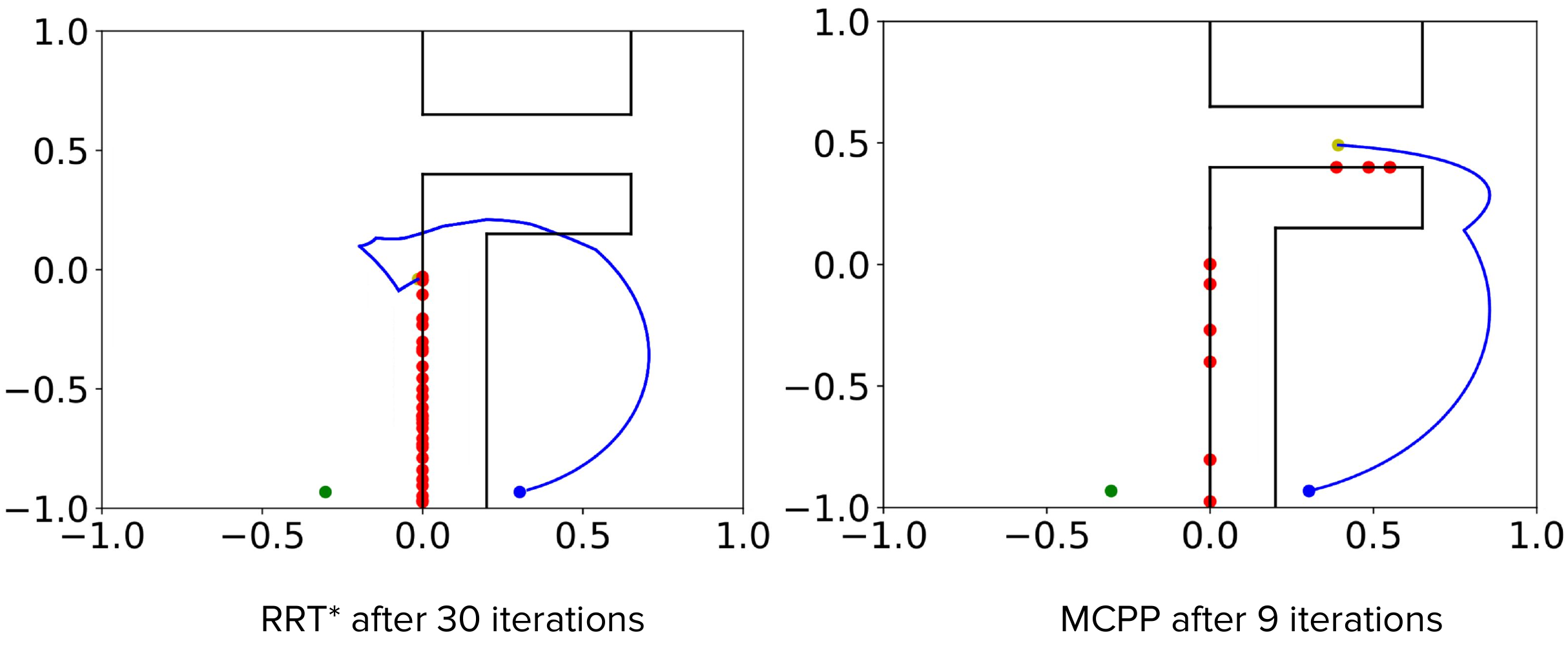}
  \vspace{-0.5em}
  \caption{L-Shape 2D POMDP. Green point is the start position. Blue point is the goal position. Red points are the collisions. The blue lines are the planning path. The figure shows a failure case of RRT* and a success case for MCPP, which shows that it is more explorative. Over 20 random seeds, RRT* failure to solve the problem with 0\% success, while MCPP obtain 100\% success with UCT and Power-UCT. TENTS gets 85\%.}  
  \label{fig:2d_reversed_l_shape}

  \centering  
  \includegraphics[scale=0.4]{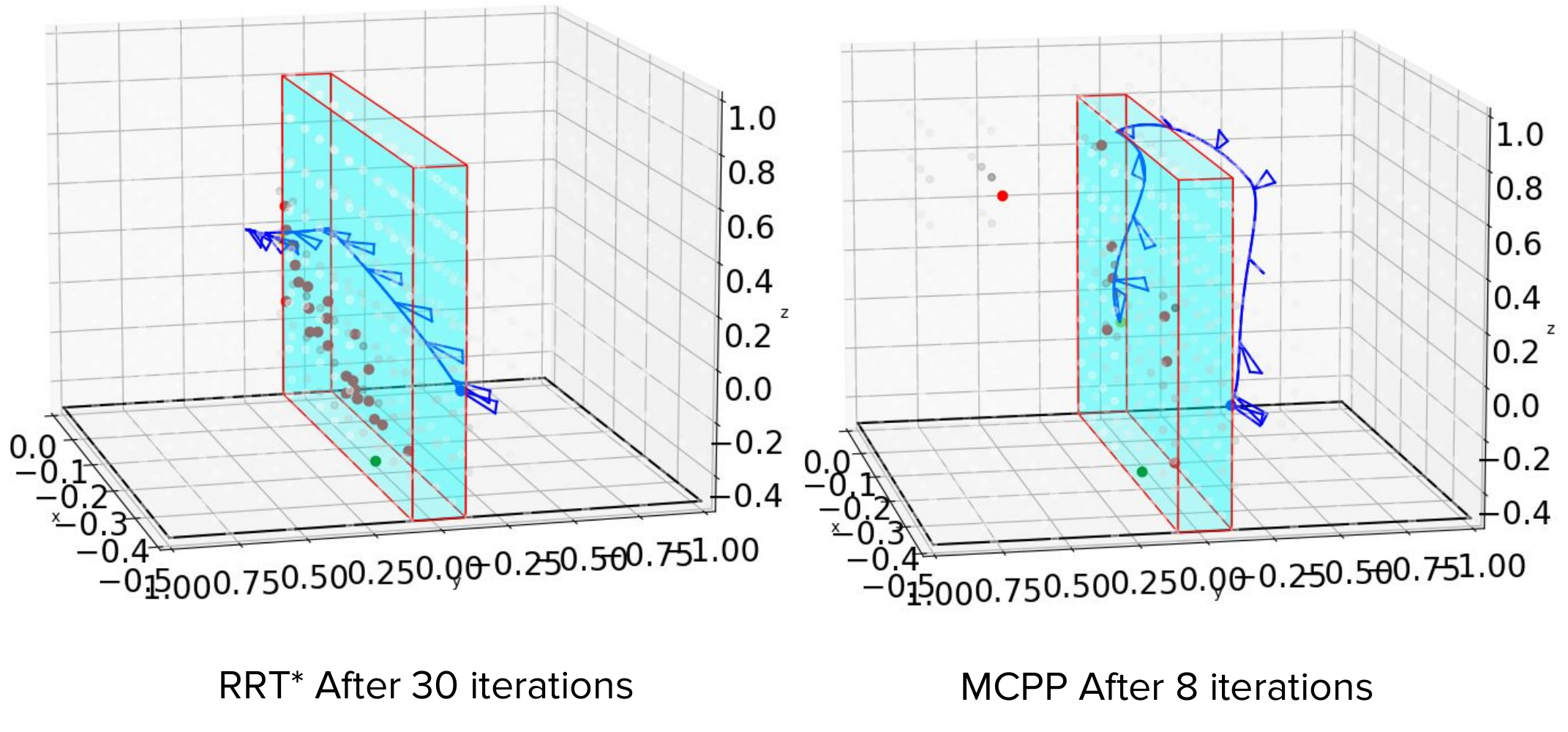}
  \caption{High Wall Environment in 3D. Grey point is the start position. Blue point is the goal position. Red points are collisions. Over 20 random seeds, $\textrm{RRT}^*$ can only success with 35$\%$, UCT obtains 55$\%$ success rate. Power-UCT gets 70$\%$ success rate while TENTS gets 45$\%$ success rate.}
  \label{fig:3d_high_wall}
  \vspace{-1.em}
\end{figure}

We provide three simulation settings in 2D and 3D state spaces to demonstrate that the MCPP planner is more explorative and can easily solve POMDP path planning problems compared to baselines. First, in a 2D U-Shape problem (Fig.~\ref{fig:2d_u_shape}), the start position is in green color while the goal position is in blue. We compare UCT, Power-UCT, and TENTS compared to $\textrm{RRT}^*$ and POMCP-DPW. As shown in Table \ref{T:pomdp_2D_u_shape}, over 100 random seeds with the same number of samples (500), UCT and Power-UCT obtain $93\%$ and $95\%$ success rate, respectively, with approximately the same number of collisions. TENTS is less explorative, with $31\%$ success rate and 18.3 collisions. POMCP-DPW gets $46\%$ success rate and 22.7 collisions while $\textrm{RRT}^*$ gets $76\%$ success rate with 14.6 collisions. The benefits of MCPP over $\textrm{RRT}^*$ can be explained as, even using a similar representation with the updating belief (probabilistic map), MCPP makes decisions based on the value function of the POMDP (by building a multistep look-ahead forward tree search), which is more explorative towards the goal. In contrast, each step decision of RRT* will be more greedy in choosing the smallest cost. Meanwhile, MCPP shows the benefit of uniformly sampling the actions inside the $\epsilon$-ball compared to POMCP-DPW, which restricts the number of actions by using the progressive widening technique. Both of the two random heuristic baselines fail to solve this task. We demonstrate one more 2D POMDP experiment with an L shape obstacle (Fig.~\ref{fig:2d_reversed_l_shape}. Over 20 random seeds with 500 samples, RRT* fails to solve the problem, while UCT and Power-UCT obtain $100\%$ success rate. TENTS is less explorative with $85\%$ success rate. POMCP-DPW obtains 85\% success rate. The two heuristic methods fail to solve this task.

\begin{table}[t!]
\scriptsize
\caption{Comparison for the U-Shape 2D POMDP}
\vspace{-1.em}
\centering
\renewcommand*{\arraystretch}{1.4}
\begin{tabular}{cccc} 
    \toprule
    {\textbf{Methods}} & $\textrm{Time(second)}$ & $\textrm{Collisions}$ & $\textrm{Success Rate}$\\
    \hline
    \midrule
 RRT* & 1555±229 & 14.6±1.5 & 76\% \\
 \cline{1-4}
 UCT & 141.7±16.3 & 15.9.±1.7 & 93.0\%\\
 \cline{1-4}
 Power-UCT & 146±17 & 15.8±1.7  & 95.0\%\\
 \cline{1-4}
 TENTS & 179.5±16 & 18.3±1.7  & 31\%\\
 \cline{1-4}
 POMCP-DPW & 322±30.8 & 22.7±1.6  & 46\%\\
\bottomrule
\end{tabular}\label{T:pomdp_2D_u_shape}
\vspace{0.5em}
\caption{Comparison for the High-Wall 3D POMDP}
\scriptsize
\centering
\renewcommand*{\arraystretch}{1.4}
\begin{tabular}{cccc} 
    \toprule
    {\textbf{Methods}} & $\textrm{Time(second)}$ & $\textrm{Collisions}$ & $\textrm{Success Rate}$\\
    \hline
    \midrule
 RRT*(bias=1) & 2854.6±4.1 & 26.0±0.4 & 10\%\\
 \cline{1-4}
 RRT*(bias=100) & 2080.3±2.6 & 19.4±1.7 & 35\%\\
 \cline{1-4}
 RRT*(bias=200) & 2548.1±2.7 & 22.2±1. & 55\%\\
 \cline{1-4}
 UCT & 178.8±13.0 & 16.4±1.0 & 55\%\\
 \cline{1-4}
 Power-UCT & 208±18.5 & 15.7±1.2  & 70.0\%\\
 \cline{1-4}
 TENTS & 267.6±17.1 & 23.0±1.2  & 45.0\%\\
 \cline{1-4}
 POMCP-DPW & 215.56±23.9 & 15.9±1.4  & 70.0\%\\
 \hline
\bottomrule
\end{tabular}\label{T:pomdp_3d_high_wall}
\vspace{0.5em}
\caption{Comparison for the real robot object disentangling}
\scriptsize
\centering
\renewcommand*{\arraystretch}{1.4}
\begin{tabular}{cccc} 
    \toprule
    {\textbf{Methods}} & $\textrm{Time(second)}$ & $\textrm{Collisions}$ & $\textrm{Success Rate}$ \\
    \hline
    \midrule
 RRT* & 1099 ±356 & 10.25 ±3 & 40\%\\
 \cline{1-4}
 UCT & 346.2 ±64 & 20.2 ±3 & 70\%\\
 \cline{1-4}
 Power-UCT & 436.5 ±177 & 22.5 ±6  & 40.0\%\\
 \cline{1-4}
 TENTS & 428.5 ±133 & 25 ±6  & 20.0\%\\
 \hline
\bottomrule
\end{tabular}\label{T:pomdp_real_robot}
\vspace{-2em}
\end{table}
Third, we build a High-wall 3D POMDP (Fig.~\ref{fig:3d_high_wall}). In this problem, the start position (green color) and the goal position (blue color) are very close, while there is a high wall standing between. The agent is not aware of the existence of the wall. As we can see in Table \ref{T:pomdp_3d_high_wall}, for 20 random seeds with the same number of samples (500), UCT obtains $55\%$ success rate with 16.4 collisions on average. Power-UCT gets a higher success rate with $70\%$ and 15.7 collisions on average. 
TENTS is less explorative with $45\%$ success rate and 23.0 collisions on average. POMCP-DPW gets $70\%$ success rate and 15.9 collisions.
On the other hand, $\textrm{RRT}^*$ can only obtain $10\%$ success rate and 26.0 collisions. Finally, the first baseline heuristic method fails to solve this task, while the second one achieves 15\% success rate.
\subsection{Real robot object disentangling task}
We compare MCPP against $\textrm{RRT}^*$ in the real-robot disentangling POMDP problem, as in~\cite{pajarinen2020probabilistic}. We use a 7-DOF KUKA LBR robot arm equipped with a SAKE gripper. Fig.~\ref{fig:disentangling_robot} illustrates the intermediate scenario of the robot arm trying to reach the goal position in the unknown box environment, while trying to disentangle the toy-bunny that was lying inside the box (the grasp part was pre-programmed). The robot does not know the shape of the box.

To evaluate the performance of our MCPP variants, we run both UCT, Power-UCT, and TENTS.
We run 10 random seeds, each random seed with 500 number of samples, and perform 30 iterations to determine if the planners can reach the goal position or not. After each iteration, if the robot hits a collision, the robot moves back a bit from the last collision position and performs the planning again with the new start position. The detailed results are shown in Table~\ref{T:pomdp_real_robot}. 
While RRT* can get 40\% success rate over ten random seeds, UCT achieves 70\% which shows the benefits of MCPP in a real-world POMDP. Power-UCT achieves 40\% success rate, and TENTS can only succeed 20\% of the times.
In terms of time, the average time in all success cases of UCT, Power-UCT, and TENTS are 364.2 seconds, 436.5 seconds, and 428.5, respectively, which are much faster compared to RRT* with an average of 1099 seconds. This can be explained as RRT* spends more time in performing the sorting to find the nearest node, as it is more biased to grow towards large unsearched areas.

\section{Conclusions}\label{sec:conclusion}
This paper addressed the major challenges of planning robot paths under partial observability from a theoretical perspective, deriving a new framework for applying MCTS planning in continuous action spaces for robot path planning. We theoretically analyzed our proposed Monte-Carlo Path Planning (MCPP) approach and proved an exponential convergence rate for MCPP for choosing the optimal path in fully observable MDPs and probabilistic completeness for finding a feasible path in POMDPs. 
Moreover, MCPP allows us to integrate different established exploration strategies in MCTS literature to improve exploration for path planning. 
We empirically analyzed our MCPP variants in benchmarks for POMDP path planning problems, showing superiority in terms of performance and computational time compared to
$\textrm{RRT}^*$, and POMCP-DPW. We further applied our new
method to a real robot POMDP problem using a KUKA 7-DOF robot arm to disentangle 
objects from other objects, without any sensory information, except for the observation of collisions. Future development involves the application of MCPP to more complicated robotic tasks and studying heuristics to accelerate the planning process with MCPP.

\bibliographystyle{IEEEtran} \bibliography{root}
\end{document}